\documentclass{article}

\PassOptionsToPackage{authoryear, round}{natbib}

\usepackage[preprint]{neurips_2024}

\setcitestyle{aysep={,}}

\usepackage[utf8]{inputenc} 
\usepackage[T1]{fontenc}    
\usepackage{hyperref}       
\usepackage{url}            
\usepackage{booktabs}       
\usepackage{amsfonts}       
\usepackage{nicefrac}       
\usepackage{microtype}      
\usepackage{xcolor}         

\usepackage{graphicx}
\usepackage{booktabs} 
\usepackage[para,online,flushleft]{threeparttable}
\usepackage{tabularx}
\usepackage{multirow}
\usepackage{multicol}
\usepackage{enumitem}
\usepackage{subcaption} 
\usepackage[nolist,printonlyused]{acronym}

\usepackage{amsmath}
\usepackage{amssymb}
\usepackage{mathtools}
\usepackage{amsthm}

\theoremstyle{plain}
\newtheorem{theorem}{Theorem}[section]

\theoremstyle{definition}
\newtheorem{definition}[theorem]{Definition}

\theoremstyle{remark}

\usepackage{pifont}

\title{Bitwidth-Specific Logarithmic Arithmetic for Future Hardware-Accelerated Training}

\author{%
  Hassan~Hamad, Yuou~Qiu, Peter~A.~Beerel, and Keith~M.~Chugg \\
  Ming Hsieh Department of Electrical and Computer Engineering \\
  University of Southern California \\
  Los Angeles, CA, USA \\
  \texttt{\{hhamad, yuouqiu, pabeerel, chugg\}@usc.edu} \\
}

\begin{document}

\begin{acronym}
\acro{LNS}{Logarithmic Number System}
\acro{CNN}{Convolutional Neural Network}
\acro{QAA-LNS}{Quantization Aware Approximate - Log Number System}
\acro{MAC}{Multiply Accumulate}
\end{acronym}

\maketitle

\begin{abstract}
While advancements in quantization have significantly reduced the computational costs of inference in deep learning, training still predominantly relies on complex floating-point arithmetic. Low-precision fixed-point training presents a compelling alternative. This work introduces a novel enhancement in low-precision logarithmic fixed-point training, geared towards future hardware accelerator designs. We propose incorporating bitwidth in the design of approximations to arithmetic operations. To this end, we introduce a new hardware-friendly, piece-wise linear approximation for logarithmic addition. Using simulated annealing, we optimize this approximation at different precision levels. A C++ bit-true simulation demonstrates training of VGG-11 and VGG-16 models on CIFAR-100 and TinyImageNet, respectively, using 12-bit integer arithmetic with minimal accuracy degradation compared to 32-bit floating-point training. Our hardware study reveals up to 32.5\% reduction in area and 53.5\% reduction in energy consumption for the proposed LNS multiply-accumulate units compared to that of linear fixed-point equivalents.~\footnote{All our code is made public at \url{https://github.com/hmhamad/QAA-LNS}}
\end{abstract}

\section{Introduction}
\label{sec:intro}

Deep learning has witnessed remarkable advancements in recent years, leading to the development of increasingly larger and more complex neural network models. However, this progress comes at a steep price – the escalating costs associated with training these neural networks. While considerable research has focused on reducing inference costs, particularly through techniques like quantization~\citep{Choi2019AccurateAE,Dettmersint8GPT}, the high computational expense of training remains a critical challenge. Training neural networks still predominantly relies on floating-point arithmetic due to its ability to handle the wide dynamic range required during the back-propagation learning pass.

In response to the rising training costs, researchers have explored methods for training neural networks using fixed-point arithmetic\footnote{While fixed-point representation is a specific form of integer arithmetic,  in the context of this work, `integer' and `fixed-point' are used interchangeably to describe the numerical format.}~\citep{Gupta2015DeepLW,wu2018WAGE,Scalable8bitBanner2018,Wang2020NITITI,Ghaffari2022IsIA}. These approaches have shown promise in reducing computational expenses but come with limitations. Integer arithmetic suffers from restricted range, precision loss, and quantization errors, introducing bias in the training process. To mitigate these issues, existing implementations often introduce additional hardware components, e.g. stochastic rounding~\citep{Gupta2015DeepLW,Ghaffari2022IsIA},  or computational overhead, e.g.  computing a per-layer scaling parameter~\citep{Wang2020NITITI}. Other studies employ specialized techniques to ensure training stability  and/or propose alternative layers or operations that are friendlier to integer arithmetic~\citep{Scalable8bitBanner2018}. Importantly, many of these studies do not perform low-bitwidth end-to-end training, often relying on full precision for sensitive operations and accumulating in higher bitwidths to prevent overflow and preserve numerical accuracy~\citep{wu2018WAGE,Scalable8bitBanner2018,Ghaffari2022IsIA}.

Amidst these efforts, fixed-point \ac{LNS} has emerged as a compelling alternative~\citep{Sanyal2019NeuralNT,Zhao2021LNSMadamLT,liu2023AlsPotq}. In \ac{LNS}, the log of the quantities of interest are used.  We  refer to the original quantities as being in the `linear domain' and the corresponding LNS quantities as being in the `log domain'. There are many potential benefits to performing fixed-point arithmetic in the log domain including the elimination of multiplies and support for a wide dynamic range similar to that of floating-point.  The elimination of multiplies in LNS is important since most computations in training are \ac{MAC} operations and multiplies dominate the circuit complexity of a \ac{MAC} operation.  While the potential of \ac{LNS} has been demonstrated to some degree, it presents its own unique challenges, particularly in addressing the addition operation. In \ac{LNS}, a  multiplication in the linear domain translates into a simple addition in the log domain, but conversely, addition in the log domain is more complicated than in the linear domain (see Section~\ref{sec:lns}). Previous studies have explored the application of the log number system to train neural networks. A notable contribution is the \ac{LNS}-Madam study~\citep{Zhao2021LNSMadamLT} in which the design of a new LNS-friendly learning algorithm, dubbed Madam, was proposed. 

Since \ac{LNS} requires  implementing an explicit approximation to the addition function, a key missing feature from previous \ac{LNS} studies is optimizing this approximation for each bitwidth (combination of integer and fractional bits). Simply speaking, for each different bitwidth format, a different approximation is used that is tailored specifically for this bitwidth. Throughout this paper, we use the terminology ``bitwidth-specific'' or ``quantization-aware'' interchangeably to refer to this idea.  To the best of our knowledge, this is the first study that explicitly incorporates bitwidth into the design of logarithmic arithmetic approximations. We first introduce a new hardware-friendly, piece-wise linear approximation to replace the complicated addition function encountered in \ac{LNS}. We constrain this approximation to use power-of-two slopes so that efficient bit-shift operations are used instead of costly multiplications. To account for the bitwidth, we employ a simulated annealing algorithm to optimize the parameters of this piece-wise linear approximation, guided by a quantization-aware loss function. We call this new format and arithmetic as ``\ac{QAA-LNS}''. This optimization process is done offline and thus does not contribute to the cost of training. Our approach will be detailed in Section~\ref{sec:qaa-lns}. To demonstrate the effectiveness of this new format, we conduct end-to-end training runs where all quantities in the network (weights, activations, gradients, and errors) are quantized-\ac{LNS} quantities and all arithmetic both in the forward and backward pass is implemented using \ac{QAA-LNS}. 

\begin{table*}
    \centering
    \caption{List of enhancements or network modifications that are common in fixed-point training literature. We view \ac{QAA-LNS} as an additional enhancement that can be combined with some of those techniques. Note that in our experiments, we avoid implementing any of those additional enhancements in order to isolate the contribution of \ac{QAA-LNS}.}
    \label{table:tricks}
    \begin{tabularx}{0.9\textwidth}{llc}
        \hline
        \textbf{Technique} & \textbf{Example Reference}  \\
        \hline
        Scaling Techniques (Loss, per-layer, per-tensor) & \citep{Wang2020NITITI} \\
        Rounding schemes (e.g. Stochastic Rounding) & \citep{Gupta2015DeepLW}  \\
        Higher bitwidth Accumulation & \citep{Ghaffari2022IsIA} \\
        Full-precision for quantization-sensitive layers/operations & \citep{wu2018WAGE}  \\
        Integer friendly layers or weight updates & \citep{Zhao2021LNSMadamLT} \\
        \hline
    \end{tabularx}
\end{table*}

The approach we present is an enhancement for low-precision training and is ``orthogonal'' to many other enhancements presented in the literature, some of which are listed in Table~\ref{table:tricks}. To demonstrate the benefits of bitwidth-specific arithmetic, our simulations do not include any additional enhancements. For instance, unlike previous studies that use higher bitwidths for accumulating intermediate values in 8-bit training, our experiments consistently use the same bitwidth for all computations. We also avoid using loss or gradient scaling, commonly employed in low-precision training, to isolate the impact of quantization-aware arithmetic. Future work will explore combining these enhancements with \ac{QAA-LNS} to potentially extend the capabilities of low-bitwidth training even further. Our contributions can be summarized as follows: 
\begin{itemize}[leftmargin=*]
    \setlength\itemsep{0em}
    \item We introduce \ac{QAA-LNS}, an approximate quantization aware log fixed-point arithmetic. The proposed approximation is hardware-friendly and optimized per bitwidth representation.
    \item A C++ bit-true simulation using \ac{QAA-LNS} successfully trains VGG and ResNet models from scratch on datasets such as CIFAR-100 and TinyImageNet using significantly lower bitwidth without sacrificing accuracy compared to that of 32-bit floating-point.
    \item A hardware study reveals up to 32.5\% reduction in area and 53.5\% reduction in energy consumption for \ac{LNS} multiply-accumulate (MAC) units when compared to their linear counterparts.
\end{itemize}

In Section~\ref{sec:back}, we look at the literature of low precision training techniques. A brief introduction to \ac{LNS} is given in Section~\ref{sec:lns} followed by a comprehensive definition of the proposed \ac{QAA-LNS} approach in Section~\ref{sec:qaa-lns}. In Section~\ref{sec:results}, we present our experimental results followed by a dedicated hardware study and ablation experiments to demonstrate the effectiveness of the quantization-aware strategy. Finally, we conclude and propose future work in Section~\ref{sec:conc}.





\section{Background and Previous Work}
\label{sec:back}
\textbf{Floating-Point Literature.} Floating-point remains the primary number system for neural network training, with \texttt{fp32} and the half-precision \texttt{fp16} format supported by modern accelerators and deep learning frameworks. Using \texttt{fp16} reduces bitwidth by half, enabling larger models and batch sizes on smaller hardware units, thus lowering energy consumption. However, \texttt{fp16}'s limited dynamic range and precision can degrade model performance, prompting the creation of variants like \texttt{bfloat16}\footnote{The bfloat16 numerical format: \url{https://cloud.google.com/tpu/docs/bfloat16}}. Research is increasingly focused on sub-16-bit quantization to address significant losses in precision and quantization errors. For example, some studies bypass or modify quantization-sensitive layers like batch-normalization or operations such as weight updates due to their failure risk at lower precision. An early study on 8-bit training introduced `Range Batch-Normalization' as a quantization-friendly alternative to traditional batch-normalization~\citep{Scalable8bitBanner2018}. In addition, 16-bit copies of the gradients were still used in the backward pass for operations not posing a performance bottleneck. Meanwhile, efforts to extend training to 4-bit precision, such as in~\citep{Sun20204bit}, have employed custom gradient scaling and specialized rounding schemes to manage reduced dynamic ranges and minimize quantization bias.\\
\textbf{Fixed-Point Literature.} Fixed-point\footnote{For a refresher on fixed-point arithmetic, refer to \emph{Fixed-Point Arithmetic: An Introduction} by Randy Yates \url{https://courses.cs.washington.edu/courses/cse467/08au/labs/l5/fp.pdf}} operations (e.g., \ac{MAC}s) can be 
implemented in circuitry that is much simpler than that required for floating-point operations, which handles the mantissa and exponent separately. This makes fixed-point particularly advantageous in resource-constrained environments because of its lower energy consumption~\citep{Horowitz2014Energy}.  Thus, fixed-point methods are widely used in many application specific signal processing circuits (e.g., communication modems). 
This potential complexity reduction has spurred a lot of interest in low-precision fixed-point arithmetic for deep learning. Quantizing neural network inference has been successfully demonstrated for extremely low bitwidths~\citep{Choi2019AccurateAE,Dettmersint8GPT}. However, its limited range and precision makes it much more challenging to utilize in training due to the wide dynamic range requirements in the backward pass. \citep{Gupta2015DeepLW} is an early study implementing 16-bit fixed-point weights for training. To mitigate the bias introduced in the rounding of fixed-point numbers, the authors proposed using a stochastic rounding scheme, which has become common in fixed-point literature. In~\citep{wu2018WAGE}, 8-bit integers were used in training but some sensitive layers and operations, such as batch-normalization and soft-max, were left in full precision floating-point. \citep{Wang2020NITITI} introduce NITI, an integer-only training framework utilizing a per-layer scaling factor for each different quantity in the network e.g., weights and gradients. Thus, the range of representable values for each quantity can be adjusted separately. NITI successfully trains with 8-bit integers but as with most previous works, accumulation is done in higher precision, in this case 32-bit. Another interesting work is that of~\citep{Ghaffari2022IsIA} where the authors manipulate the floating-point format to switch it into 8-bit integers instead of using quantization directly. Still, accumulation is done in higher bitwidth and non-linear operations aside from the ReLu activation remain in floating-point.\\
\textbf{\ac{LNS} Literature.} \ac{LNS} traces its origins to foundational works such as~\citep{SwAl75},~\citep{KiNiRa71} and~\citep{LeEdAl77}. One of the main drawbacks of fixed-point formats is the limited range due to the resolution being fixed. A fixed step-size in the linear domain becomes a variable step-size in the log domain because of the non-linearity of the $\log$ function. Therefore, in \ac{LNS}, performing the quantization in the log domain produces a variable resolution and an exponentially larger representable range. A representable range of $[c,d]$ in the linear domain becomes $[b^c, b^d]$ in the log domain, where $b$ is the logarithm base. While LNS offers significant advantages in terms of range and resolution, it also has potential limitations. In \ac{LNS}, the non-uniform distribution of quantization levels leads to larger spacing between levels for higher magnitudes. This can result in higher quantization errors for large values compared to fixed-point representations with uniform spacing. However, it is important to note that the impact of these quantization errors on the overall performance of \ac{LNS}-based systems depends on the specific application and the range of values encountered. To mitigate the potential impact of quantization errors, careful selection of the LNS parameters, such as the number of fractional bits and the logarithm base, is crucial. Additionally, multiplications transform into additions because of the $log$ property: $\log(ab) = \log(a) + \log(b)$. The downside is that the addition operation $\log(a+b)$  has no simple form in the log domain. This drives most of the research on \ac{LNS} to find an efficient approximation for addition~\citep{Arnold2020LNS}. \citep{ArBaCuWi97} is an early study which noticed significant hardware savings when implementing logarithmic arithmetic in back-propagation. In \citep{Sanyal2019NeuralNT}, a simple bit-shift approximation for addition is utilized to train small networks using 16-bit \ac{LNS} on MNIST-like datasets while accumulating in 32-bit. Our experiments show that a much more precise approximation is needed for large-scale experiments. The closest method to ours is~\citep{Zhao2021LNSMadamLT} where LNS-Madam is introduced to successfully train using 8-bit \ac{LNS}, with 24-bit accumulation, on large-scale datasets such as ImageNet~\citep{Deng2009ImageNet}. To achieve this result, a multiplicative weight update scheme is introduced, dubbed Madam, as an alternative to Adam where the weight updates are additive. A key missing idea from these previous \ac{LNS} studies is optimizing the addition approximation function for each specific bitwidth configuration. In simple terms, for each combination of integer and fractional bits in the \ac{LNS} format, a different and unique approximation for addition should be used that is tailored for this bitwidth configuration. In this paper, we sometimes use ``bitwidth-specific'' or ``quantization-aware'' to describe this concept. To the best of our knowledge, this is the first study that has designed logarithmic arithmetic approximations with bitwidth in mind.

\section{LNS Definition and Arithmetic}\label{sec:lns}
\begin{figure*}
    \centering
    \begin{subfigure}[b]{0.49\columnwidth}
        \includegraphics[width=\textwidth]{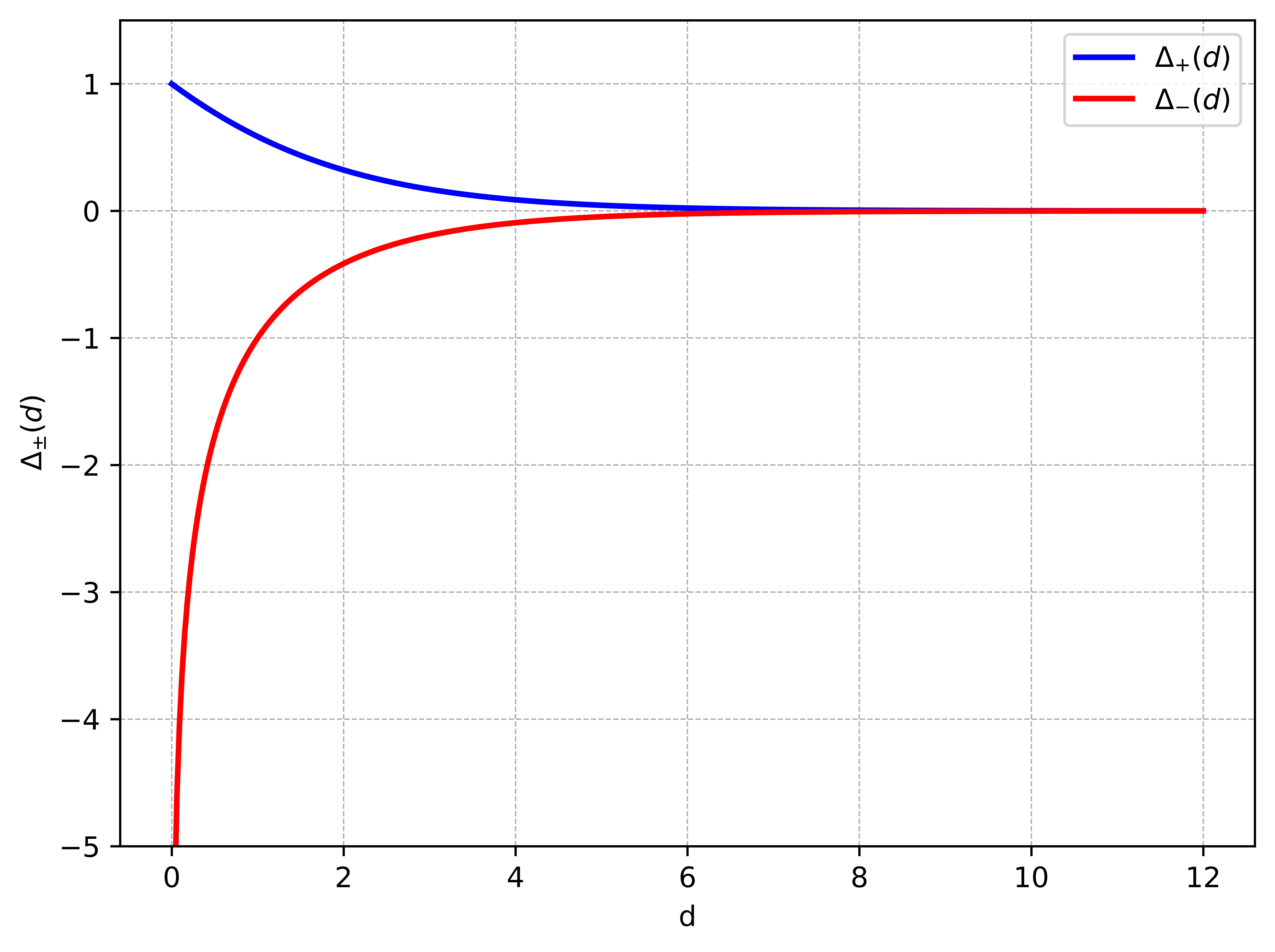}
        \caption{Plot of the $\Delta_{+}(d)$ and $\Delta_{-}(d)$ functions over the range $d \in [0,12]$.}
        \label{fig:sub1}
    \end{subfigure}
    \hfill 
    \begin{subfigure}[b]{0.49\columnwidth}
        \includegraphics[width=\textwidth]{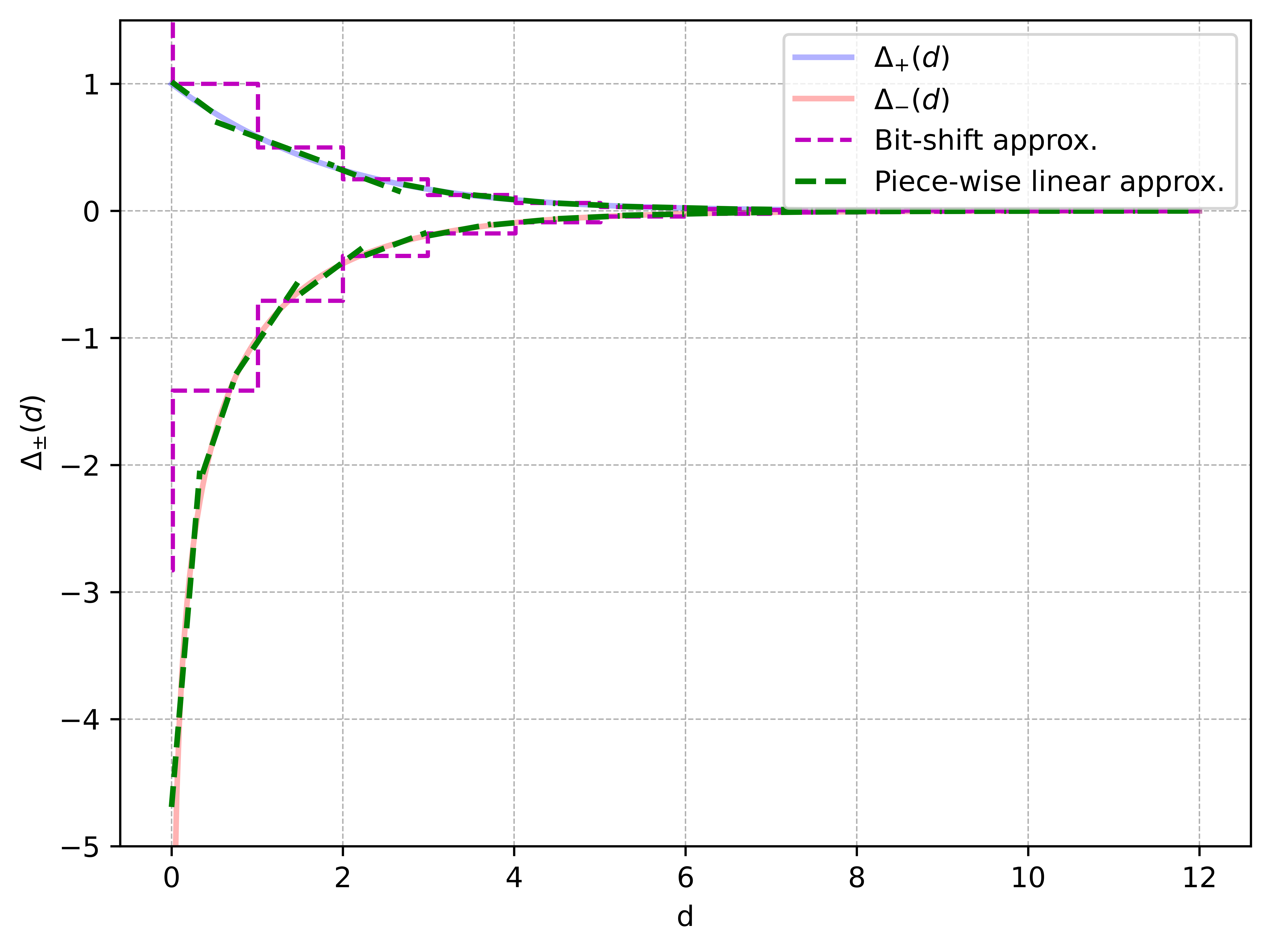}
        \caption{An example of a bit-shift approximation and a 16-segment piece-wise linear approximation..}
        \label{fig:sub2}
    \end{subfigure}
    \caption{Plot of the $\Delta_{+}(d)$ and $\Delta_{-}(d)$ functions (a) with an example of a bit-shift approximation and our proposed piece-wise linear approximation with power-of-two slopes (b). Notice how both $\Delta_{+}(d)$ and $\Delta_{-}(d)$ functions converge to zero within a small input range. This allows us to design an approximation over a small range without increasing quantization error. Also notice in (b) how the piece-wise linear segments provide a more accurate approximation to the curves.}
    \label{fig:delta_functions}
\end{figure*}

In this section, we formalize the definition of \ac{LNS} and describe the arithmetic operations within this system. A real number $x$ in the `linear domain' signifying a quantity in the training process, e.g. a weight or a gradient, can be equivalently represented in the `log domain' by a tuple as follows:
\begin{equation}\label{eq:lns}
    x \leftrightarrow (\log_2(|x|), s_x)
\end{equation}
 where $s_x$ represents the sign of $x$. Next, we expand this representation to account for quantization.
\begin{definition}[\ac{LNS} Fixed-Point Representation]
Let $x$ be a real number and let $T$, $I$, and $F$ denote the total number of bits, integer bits, and fractional bits, respectively, used in the log magnitude representation of $x$ such that $T = I + F$. The fixed-point \ac{LNS} representation of $x$ is expressed as a tuple $(\ell_x,s_x)$ where the sign flag $s_x$ is equal to one for negative $x$ and zero for non-negative $x$ and the log-magnitude is 
\begin{equation}\label{eq:lns-mag}
\ell_x = \text{clip}\left(\text{round}\left(\log_{2}(|x|) \times 2^{F}\right), -2^{2^I-1} -1, 2^{2^I-1}\right)
\end{equation}
where $\text{clip}(v, \text{min}, \text{max})$ bounds $v$ within the range specified by $\text{min}$ and $\text{max}$, and $2^F$ serves as the scaling factor, locating the binary point at the desired fixed position. In total, $T+1$ bits are needed to represent $x$ in the log domain including the single bit for the sign $s_x$.
\end{definition}
Multiplication in \ac{LNS} is straightforward, as it translates to addition of the logarithmic components. Addition in \ac{LNS}, however, is more involved due to the non-linear nature of the logarithmic mapping.
\begin{definition}[\ac{LNS} Multiplication]
Given two \ac{LNS} numbers $(\ell_x,s_x)$ and $(\ell_y,s_y)$, the linear product $z = x * y$ is defined in \ac{LNS} as:
\begin{equation}\label{eq:lns-mult}
(\ell_z,s_z) = (\ell_x + \ell_y,s_x \oplus s_y)
\end{equation}
where $\oplus$ denotes the XOR operation on the sign bits, yielding the sign of the product.
\end{definition}
\begin{definition}[\ac{LNS} Addition]
Given two \ac{LNS} numbers $(\ell_x,s_x)$ and $(\ell_y,s_y)$. The linear sum $z = x + y$ is defined in \ac{LNS} as $(\ell_z,s_z)$ where:
\begin{equation}\label{eq:lns-add-mag}
\begin{aligned}
\ell_z &= \log_2(|x+y|) = \log_2(|s_x \cdot 2^{\ell_x} + s_y \cdot 2^{\ell_y}|) \\
        &= \begin{cases} 
            \max(\ell_x,\ell_y) + \Delta_{+}(|\ell_x-\ell_y|) & \text{if } s_x = s_y \\
            \max(\ell_x,\ell_y) + \Delta_{-}(|\ell_x-\ell_y|) & \text{if } s_x \ne s_y
          \end{cases}
\end{aligned}
\end{equation}
\begin{align}\label{eq:deltas}
    \text{with } \Delta_{+}(d) &= \log_2(1+2^{-d}) & \text{for } d \geq 0 \quad \text{ and } \quad
    \Delta_{-}(d) &= \log_2(1-2^{-d}) & \text{for } d \geq 0
\end{align}
where $\Delta{\pm}$ are defined for ease of notation. The sign bit for the sum is given by $s_z = s_x~\text{ if } \ell_x \geq \ell_y ~\text{ and } s_y~\text{ if } \ell_x < \ell_y $.
\end{definition}
\textbf{Approximations.} Figure~\ref{fig:delta_functions} (a) depicts the $\Delta_{+}(d)$ and $\Delta_{-}(d)$ curves. Because it is not possible to implement the functions in (\ref{eq:deltas}) exactly in fixed-point, i.e. using integer math, researchers target efficient approximations. This is particularly viable for machine learning applications due to the inherent tolerance to noise, as long as the introduced numerical errors are unbiased. For instance, in~\citep{Sanyal2019NeuralNT}, a simple bit-shift approximation method, as illustrated in Figure~\ref{fig:delta_functions} (b), is applied to relatively small-scale experiments. While cost-effective, our experiments reveal that more precise approximations are needed for training large-scale models. This is also the case in~\citep{Arnold2020LNS}, where the Mitchell's approximation~\citep{Mitchell1962} for the logarithm is used to implement back-propagation in \ac{LNS} for small networks. In~\citep{Zhao2021LNSMadamLT} a hybrid approach, combining Mitchell's approximation with table lookup and subsequent bit-shift operations was adopted to train on large-scale datasets. This design, coupled with other enhancements, enables successful 8-bit LNS training, although accumulation still relies on 24-bit precision. 

\section{Quantization Aware Approximate - Log Number System (QAA-LNS)}\label{sec:qaa-lns}
In this section, we introduce the proposed Quantization Aware Approximate - Log Number System (QAA-LNS) approach. This method differs from previous approximation techniques by accounting for quantization in the $\Delta_{\pm}$ approximations. \\ 
\textbf{Efficiency.} Our design uses a piece-wise linear approximation featuring power-of-two slopes, which enables computationally expensive multiplications to be replaced by more efficient bit-shift operations. This piece-wise linear curve, as shown in Figure~\ref{fig:delta_functions} (b), is defined by  selection of the bin locations, slopes, and offset values for each segment of the curve. The resulting approximation can be expressed as:
\begin{equation}\label{eq:approx}
    \Delta(d) \approx d \times a_i + o_i = d \times 2^{k_i} + o_i = \text{Bit-shift}(d,a_i) + o_i
\end{equation}
    where $i$ is the index of the bin $c_i$ which input $d$ falls into, $a_i = 2^{k_i}$ and $o_i$ are the slope and offset associated with segment $i$, respectively, and $k_i \in \mathcal{Z}$. This approximation is used for both $\Delta_{+}$ and $\Delta_{-}$ functions with different bin boundaries, slopes and offsets for each.\\
\textbf{Accuracy.} To obtain an accurate approximation, the bin locations, slopes, and offsets must be optimized. Towards this end,  consider two extremes: On the one end, the straightforward method could be to simply minimize the mean squared error (MSE) between the true curve $\Delta_{\pm}$ and the approximate curve, call it $\hat{\Delta}_{\pm}$, at some $N$ sample points $d_n,~ n=1,..,N$ as follows:
\begin{equation}\label{eq:loss-not-qa}
    L = \frac{1}{N} \sum_{n=1}^{N} (\Delta_{\pm}(d_n) - \hat{\Delta}_{\pm}(d_n))^2 
\end{equation}
While this is straightforward to implement, it does not account for quantization. Our experiments in Section~\ref{sub:ablation} show how training diverges when applying this `non quantization-aware' approach. On the other extreme, one could conceptually consider optimizing the approximation parameters at a given fixed-point format via end-to-end learning performance. While this accounts for quantization, it is likely too computationally complex and is optimized only for a particular dataset and model.

In \ac{QAA-LNS}, we consider a middle-ground approach. The selection of bin locations, slopes, and offsets is optimized for a general \ac{LNS} addition operation at a particular bitwidth. Consider two real vectors $\underline{x}$ and $\underline{y}$ in the linear domain, e.g. weights or activations. Denote the exact sum as $\underline{z} = \underline{x} + \underline{y}$. Now define $\tilde{\underline{z}}$ to be the quantized value of $\underline{z}$ in the linear domain, i.e. quantize $\underline{z}$ to \ac{LNS} as in (\ref{eq:lns-mag}) then convert back to linear domain with infinite precision. Thus, $\tilde{\underline{z}}$ represents the ideal values we would like to obtain when performing \ac{LNS} addition of $\underline{x}$ and $\underline{y}$ . Next, denote $\hat{\underline{z}}$ to be the approximate output obtained from \ac{LNS} addition in (\ref{eq:lns-add-mag}). Define the loss function as:
\begin{equation}\label{eq:loss}
    L(\underline{x},\underline{y}) = \frac{1}{N} \sum_{n=1}^{N} (\tilde{z}_n - \hat{z}_n)^2
\end{equation}
where $\tilde{\underline{z}}$ and $\hat{\underline{z}}$ are implicit functions of $(\underline{x},\underline{y})$, the bitwidths, and the delta approximation parameters.  
Finally, we employ the simulated annealing algorithm~\citep{Kirkpatrick1983SA} with a cosine cooling schedule to perform the optimization over the approximation parameters. To generate the neighbor solution at time $t$ in the algorithm, a random bin $c_i(t)$ is selected, and the new bin location $c_i(t+1)$ is then sampled from a uniform distribution in the interval $[c_{i-1}(t),c_{i+1}(t)]$. The algorithm then iterates over a finite list of power-of-two slopes and finds the combination of slope $a_i(t+1)$ and offset $o_i(t+1)$ which minimize the mean squared-error in the segment $i$.\\
\textbf{Discussion.} By incorporating quantization into our loss function, the resulting approximation is tailored to the intricacies of each bitwidth representation. As we demonstrate in the subsequent experiments in Section~\ref{sub:ablation}, this aspect proves to be crucial for achieving successful convergence during training. For instance, a visual inspection of the approximation curves (comprising bin locations, slopes, and offsets) for 11-bit \ac{LNS} and 14-bit \ac{LNS} may not reveal any visible difference. However, these seemingly subtle distinctions arising from simulated annealing are essential in avoiding the accumulation of numerical errors throughout the training process, preventing divergence.\\
\textbf{Selection of Parameters.} For our experimental setup, we employ the following parameters: $N=10,000$, $\underline{x}$ and $\underline{y}$ sampled from a $\mathcal{N}(0,3)$ distribution, and a $16$-segment piece-wise linear curve over the range $d \in [0,12]$. The normal distribution is a natural choice for the sample vectors $\underline{x}$ and $\underline{y}$ since this resembles the distributions of weights and activations in the network. Nevertheless, multiple distributions were tested and all were found similarly effective. Similarly, the number of segments/bins was varied: specifically $8, 12, 16$, and $32$, were tested and the size of $16$ was found to provide the best balance between complexity and accuracy. Regarding the range, an inspection of Figure~\ref{fig:delta_functions} reveals how both curves tend to zero for large $d$. Around $d=12$, the values of $\Delta_{+}$ and $\Delta_{-}$ start to get below the resolution for most low bitwidth precision. 

\begin{table*}
    \centering
    \caption{Median of validation accuracy for the last 10 training epochs of 32-bit floating-point (baseline) vs fixed-point \ac{LNS} using the VGG Models. The \ac{LNS} bitwidth representation, $T$-bit $(F,o)$ emphasizes that \ac{LNS} arithmetic units such as addition and multiplication act on $T$-bit integers with $F$ dedicated fractional bits. $o$ is the number of overhead flags used in the representation.}
    \label{table:results-vgg}
    \small
    \begin{tabular}{l|l||c|c||c|c}
        \toprule
        \multirow{2}{*}{\textbf{Format}} & \multirow{2}{*}{\textbf{Bitwidth}} & \multicolumn{2}{c||}{\textbf{VGG-11 on CIFAR-100}} & \multicolumn{2}{c}{\textbf{VGG-16 on TinyImageNet}} \\
        \cline{3-6}
        & &  \textbf{Acc} & \textbf{Degradation} & \textbf{Acc} & \textbf{Degradation} \\
        \midrule 
        Floating-Point & 32-bit & 66.03\% ± 0.14\% & - & \textbf{54.47\% ± 0.31\%}& - \\
        \midrule
        \multirow{3}{*}{QAA-LNS} & 14-bit (F=8, o=2) & \textbf{66.13\% ± 0.24\%} & -0.1\% & 53.65\% ± 0.88\% & 0.82\% \\
         & 12-bit (F=6, o=2) & 65.89\% ± 0.45\%  & 0.14\% & 53.60\% ± 0.76\% & 0.87\% \\
         & 11-bit (F=5, o=2) & 59.29\% ± 4.77\% & 6.74\% & 37.51\% ± 3.06\% & 16.96\% \\
        \bottomrule
    \end{tabular}
\end{table*}

\section{Experimental Results}\label{sec:results}
To demonstrate the value of \ac{QAA-LNS}, we conduct several image classification experiments. This choice was driven by two main reasons: first, they come with well-established benchmarks that enable reliable comparative analysis; second, the diverse layer types found in convolutional neural networks (CNNs), such as convolutional layers, batch normalization, and max pooling, allow us to demonstrate the successful application of \ac{LNS} arithmetic across different neural network components.

\begin{figure}
\centering
\includegraphics[width=0.5\columnwidth]{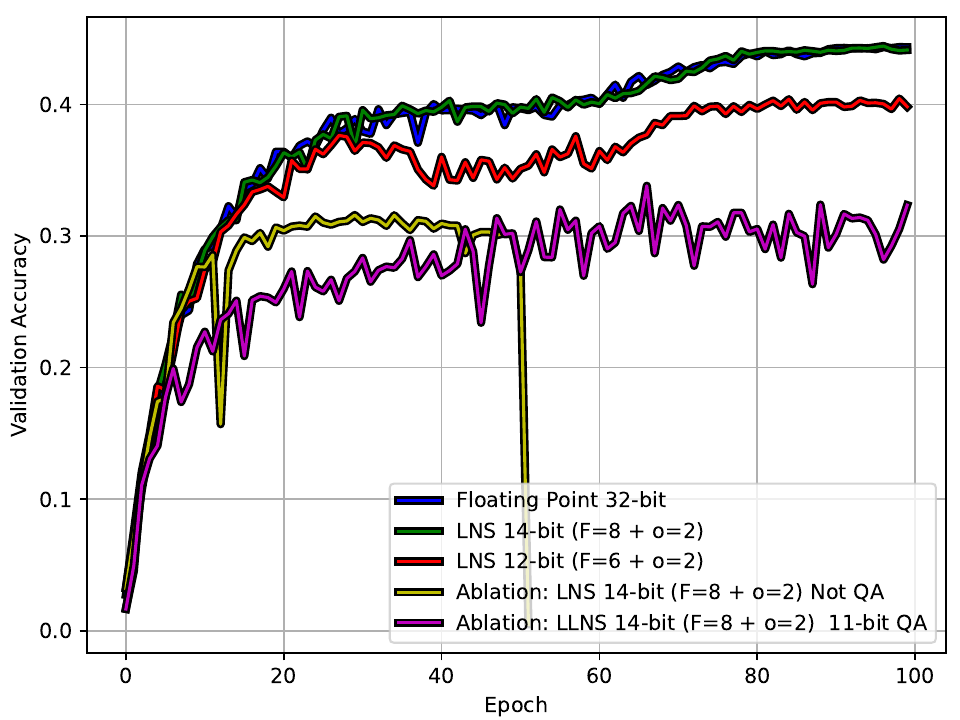}
\caption{Validation accuracy curves with ResNet-18 on TinyImageNet for multiple experiments. To reach the floating-point level, at least $14$ `arithmetic' bits are needed. $12$-bit \ac{LNS} suffers from severe degradation in accuracy. `Not QA` refers to using an approximation that is not `quantization-aware'. `$11$-bit QA' refers to using a `quantization-aware' approximation that was optimized for $11$-bit \ac{LNS}.}
\label{fig:learning_curves}
\end{figure}

\textbf{Datasets.} We evaluate our method on two popular image datasets: CIFAR-100~\citep{Krizhevsky09learningmultiple} and TinyImageNet~\citep{TinyImageNet2017}. CIFAR-100 contains 60,000 training images and 10,000 validation images across 100 classes. TinyImageNet, a scaled-down version of the larger ImageNet dataset~\citep{Deng2009ImageNet}, includes 100,000 training images and 10,000 validation images from 200 classes, with images resized to $64\times 64$ pixels.\\
\textbf{Models and Setup.} We use two popular CNN architectures, VGG~\citep{SimonyanZ14a} and ResNet~\citep{He1025ResNet}, for our experiments. For TinyImageNet, we train ResNet-18 and VGG-16 models, while for CIFAR-100, we train VGG-11 and the CIFAR version of ResNet-18 (with a $3 \times 3$ convolution replacing the initial $7 \times 7$ convolution and removing the MaxPooling layer to better suit the CIFAR-100 image scale). All layers and operations are implemented in fixed-point \ac{LNS}. We use RandomFlip and RandomCrop augmentations with $4$ pixel padding for $32 \times 32$ resolution on CIFAR-100 and $56 \times 56$ on TinyImageNet. A hyper-parameter search was performed using floating-point and the same settings were then used for \ac{LNS}: SGD with weight decay of $0.0001$, momentum of $0.9$, batch size of $128$, and a Cosine Scheduler with Warm Restarts~\citep{Loshchilov2016SGDR}, starting at learning rates of $0.1$ for CIFAR-100 and $0.5$ for TinyImageNet. Each model was trained for $100$ epochs only. Extended training in \ac{LNS} is computationally intensive.\\
\textbf{\ac{LNS} Setting}. In our training configurations, we included an additional `zero-flag' overhead bit in our \ac{LNS} format to handle zero values, as the logarithm of zero is undefined. Alternatively, we explored representing zero by the smallest possible value within the bitwidth, which Section~\ref{sub:ablation} shows can eliminate the need for a zero-flag bit without loss of accuracy. In addition, the soft-max function in the final layer requires computing an exponential. Since this is only used once in the final layer, not posing a performance bottleneck, we utilize a relatively large piece-wise linear approximation, similar to that introduced in Section~\ref{sec:qaa-lns}, consisting of $256$-segments. We tested three \ac{LNS} bitwidth configurations, all including $o=2$ flag bits (one for sign and one for zero-flag): $T=14$-bit with $F=8$ fractional bits, $T=12$-bit with $F=6$ fractional bits, and $T=11$-bit with $F=5$ fractional bits. We refer to  $T$ as the arithmetic bitwidth since, for example, the addition $(\ell_x + \ell_y)$ in (\ref{eq:lns-mult}) is performed by a standard two's complement $T$ bit adder. In other words, the overhead bits have negligible effect on the LNS \ac{MAC} circuit complexity and we therefore refer to these schemes as `$T$-bit QAA-LNS.' Our objective is to present cases where training reaches floating-point levels as well as those leading to significant degradation in accuracy, i.e. failure cases.\\
\textbf{Evaluation.} In each experimental configuration, we conduct two training runs utilizing different random seeds and report the mean and standard deviation of the validation accuracy. We used the median of the validation accuracy from the last $10$ epochs as our evaluation metric. All experiments were carried out on NVIDIA RTX 2080 Ti or RTX 3090 GPUs without using any deep learning frameworks. Instead, we write custom CUDA kernels to implement \ac{LNS} on the GPUs.\\
\textbf{Results.} The results from our experiments with VGG and ResNet models are presented in Tables~\ref{table:results-vgg} and~\ref{table:results-resnet}, respectively. We also depict validation accuracy curves from our experiments in Figure~\ref{fig:learning_curves}. Using VGG models, training on CIFAR-100 and TinyImageNet with 12-bit arithmetic shows negligible accuracy degradation compared to the 32-bit floating-point baseline. Accuracy declines when using 11-bit arithmetic. For ResNet, similar results are seen on CIFAR-100, while TinyImageNet required 14-bit arithmetic to maintain accuracy. On both datasets, training with lower bitwidths led to divergence, highlighting that larger models and datasets need higher precision for effective training.\\
\textbf{Software Methods, Limitations, and Comparisons.} As current software libraries and hardware do not support \ac{LNS} fixed-point format and arithmetic, we had to develop our \ac{LNS} code setup from scratch in C++ and CUDA, without the use of any deep learning framework. While \ac{LNS} fixed-point is proposed as a cost-effective arithmetic format, it requires specialized hardware to perform operations like those in (\ref{eq:lns-mult}), (\ref{eq:lns-add-mag}), and (\ref{eq:approx}). Even with our \ac{LNS} CUDA kernels, \ac{LNS} operations cannot fully utilize GPU CUDA cores, which are optimized for linear multiply-accumulate tasks. For instance, training a ResNet-18 on CIFAR-100 with \ac{LNS} takes over an hour per epoch. Consequently, training on larger datasets or for longer periods becomes computationally prohibitive. Since much of the low-bitwidth training literature involves ImageNet or prolonged training epochs, a direct comparison with the literature is difficult. However, our focus is mainly to highlight the benefits of bitwidth-specific arithmetic. Results discussed here and in Section~\ref{sub:ablation} illustrate the superiority of this approach over non-bitwidth-specific designs. The goal of this paper is not to achieve the state of the art in minimizing the training bitwidth. In fact, prior studies have already demonstrated successful training with 8-bit representations~\citep{Wang2020NITITI,Scalable8bitBanner2018,Zhao2021LNSMadamLT, Ghaffari2022IsIA} utilizing the various enhancements in Table~\ref{table:tricks}. We consider our work as a new enhancement that can be done \textbf{offline} with no added cost to training. A natural direction for future research is pushing the limits of bitwidth reduction in \ac{LNS} by utilizing similar techniques to those in Table~\ref{table:tricks}.

\begin{table*}
    \centering
    \caption{Median of validation accuracy for the last 10 training epochs of 32-bit floating-point (baseline) vs fixed-point \ac{LNS} using the ResNet-18 models.}
    \label{table:results-resnet}
    \small
    \begin{tabular}{l|l||c|c||c|c}
        \toprule
        \multirow{2}{*}{\textbf{Format}} & \multirow{2}{*}{\textbf{Bitwidth}} & \multicolumn{2}{c||}{\textbf{ResNet-18 on CIFAR-100}} & \multicolumn{2}{c}{\textbf{ResNet-18 on TinyImageNet}} \\
        \cline{3-6}
        & &  \textbf{Mean ± Std} & \textbf{Degradation} & \textbf{Mean ± Std} & \textbf{Degradation} \\
        \midrule 
        Floating-Point & 32-bit & \textbf{71.83\% ± 0.22\%} & - & \textbf{44.53\% ± 0.25\%} & - \\
        \midrule
        \multirow{3}{*}{QAA-LNS} & 14-bit (F=8, o=2) & 71.49\% ± 0.15\% & 0.34\% & 43.92\% ± 0.30\% & 0.61\% \\
         & 12-bit (F=6, o=2) & 71.63\% ± 0.30\% & 0.20\% & 38.79\% ± 1.30\% & 5.74\% \\
         & 11-bit (F=5, o=2) & 65.00\% ± 3.48\% & 6.83\% & 35.39\% ± 0.16\% & 9.15\% \\
        \bottomrule
    \end{tabular}
\end{table*}

\begin{table*}
    \centering
    \caption{Area and power consumption for a MAC based on different number systems. The bitwidth listed below refers to the total number of bits used by any number representation. For QAA-LNS, this is $T+o$ where $T$ is the number of arithmetic bits and $o=2$ is the number of overhead flag-bits.}
    \label{table:hardware}
    \small
    \begin{tabular}{|l||c|c|c|}
        \hline
        \textbf{Area ($\mu$m\(^2\)) / Power ($\mu$W)} & \textbf{QAA-LNS MAC} & \textbf{Fixed-Point INT MAC} & \textbf{ Floating-Point MAC} \\
        \hline
        14-bit & 2051.7 / 135.6 ~($T=12$) & 2736.4 / 237.7 & - \\
        \hline
        16-bit & 2599.9 / 173.3 ~($T=14$) & 3494.8 / 318.2 & 5259.9 / 400.6 \\
        \hline
        18-bit & 3215.1 / 220.0 ~($T=16$) & 4341.9 / 413.8 & - \\
        \hline
        20-bit & 3745.9 / 263.7 ~($T=18$) & 5280.5 / 522.1 & - \\
        \hline
        22-bit & 4258.4 / 301.6 ~($T=20$) & 6309.2 / 648.7 & - \\
        \hline
    \end{tabular}
\end{table*}

\subsection{Hardware Study}\label{sub:hardware}
We implement QAA-\ac{LNS}-based \ac{MAC}s with different bitwidth configurations at the RTL level. Linear fixed-point (integer) MACs as well as floating-point MACs are taken as baseline designs. Given the large dynamic range of LNS and floating-point formats, overflow is uncommon in training. Consequently, we set the outputs of \ac{LNS} MACs and the floating-point MACs to the same bitwidth as the inputs. For the INT MACs we followed the standard practice of using the multiplier output bitwidth of twice the input bitwidth to avoid overflow.  Table~\ref{table:hardware} summarizes the area and power consumption of the proposed QAA-LNS MACs and baseline designs as a function of input bitwidth synthesised using an open source free 45nm cell library at 1.0V and 100MHz. The \ac{LNS} MACs demonstrate substantial benefits, saving up to 32.5\% in area and 53.5\% in power, over INT MACs for the same input bitwidth. This advantage diminishes with decreasing bitwidth. Nevertheless, even in a $16$-bit configuration, the \ac{LNS} MAC exhibits noteworthy benefits, saving up to 25.6\% in area and 45.5\% in power, compared to the INT counterpart, and saving 50.5\% in area and 56.7\% in power compared to the FP16 counterpart. Based on our training experiments and the results from \citep{Gupta2015DeepLW}, it is reasonable to compare QAA-LNS at 16-bit ($T=14$) to the linear \ac{MAC} at 20-bit, which is a reduction in power and area of 66.8\% and 50.7\%, respectively.  These results suggest a compelling case for significant overall area and energy benefits for QAA-LNS in neural network training. 
Lastly, we note that almost half of the logic in the QAA-LNS MAC is required to determine the bin to which the input belongs during the computation of the approximation for $\Delta_{\pm}$. Further optimization of these boundaries to be more hardware-friendly may yield further benefits.

\begin{table}
    \centering
    \caption{Ablation Studies using ResNet-18 on CIFAR-100 and TinyImageNet}
    \label{table:ablation}
    \small
    \begin{tabular}{|l|c|c|}
        \hline
        \textbf{Study} & \textbf{Dataset} & \textbf{ Accuracy} \\
        \hline
        \hline
        FP 32-bit & TinyImageNet & 44.53\% \\
        \hline
        LNS 14-bit (Not QA) & TinyImageNet & 30.40\% \\
        \hline
        LNS 14-bit (11-bit QA) & TinyImageNet & 29.09\% \\
        \hline
        \hline
        FP 32-bit & CIFAR-100 & 71.83\% \\
        \hline
        LNS 14-bit (No zero-flag) & CIFAR-100 & 72.01\% \\
        \hline
    \end{tabular}
\end{table}

\subsection{Ablation Studies}\label{sub:ablation}
Two ablation experiments were conducted to demonstrate the effectiveness of the proposed `quantization-aware' approximation. Using ResNet-18 on TinyImageNet: Run $T=14$-bit \ac{LNS} with an approximation that does not incorporate quantization, i.e. approximating the $\Delta_{\pm}$ curves directly as in (\ref{eq:loss-not-qa}). Next, use the approximation optimized for $T=11$-bit \ac{LNS} for training with $T=14$-bit representation. Results are displayed in Table~\ref{table:ablation} and the learning curves are depicted in Figure~\ref{fig:learning_curves}. A severe degradation compared to the floating-point baseline can be seen in both cases. In the case of using the non quantization-aware approximation, the gradients of the network blow up in magnitude and cause numerical instability. These two results not only show that incorporating quantization effects is crucial for ensuring numerical stability but also emphasize that the approximation must be tailored for each specific bitwidth separately. Additionally, a third experiment with ResNet-18 on CIFAR-100 employed $T=14$-bit \ac{LNS} without a zero-flag bit, demonstrating that omitting the zero-flag bit (using $F=8$ fractional bits + $o=1$ flag bit) had no negative impact in this context and can be dropped, although further investigation is needed for lower bitwidths. 
\vspace{-0.5em}
\section{Conclusion and Future Work}\label{sec:conc}
This study has demonstrated the value of the bitwidth-specific arithmetic in enhancing low-precision neural network training. We have shown that it is possible to achieve near floating-point accuracy with significantly reduced computational overhead. There are many fruitful directions for future work. Alternative approximation structures for LNS addition can be explored using the quantization-aware optimization approach. We avoided mixed-precision and non-standard modifications to the training process, but future work could apply such methods to aggressively minimize the required bitwidths for LNS training. A significant scaling up of end-to-end training with LNS was achieved, but slow run-times remain a limitation. This could be potentially addressed by further CUDA optimizations or direct hardware implementation (e.g., in FPGA). Finally, extension to transformer networks is an interesting direction that would require a more aggressive optimization for soft-max.  

\bibliographystyle{plainnat}
\bibliography{references}

\begin{thebibliography}{25}
\providecommand{\natexlab}[1]{#1}
\providecommand{\url}[1]{\texttt{#1}}
\expandafter\ifx\csname urlstyle\endcsname\relax
  \providecommand{\doi}[1]{doi: #1}\else
  \providecommand{\doi}{doi: \begingroup \urlstyle{rm}\Url}\fi

\bibitem[Arnold et~al.(2020)Arnold, Chester, and Johnson]{Arnold2020LNS}
Mark Arnold, Ed~Chester, and Corey Johnson.
\newblock Training neural nets using only an approximate tableless lns alu.
\newblock In \emph{2020 IEEE 31st International Conference on Application-specific Systems, Architectures and Processors (ASAP)}, pages 69--72, 2020.
\newblock \doi{10.1109/ASAP49362.2020.00020}.

\bibitem[Arnold et~al.(1997)Arnold, Bailey, Cupal, and Winkel]{ArBaCuWi97}
MG~Arnold, TA~Bailey, JJ~Cupal, and MD~Winkel.
\newblock On the cost effectiveness of logarithmic arithmetic for backpropagation training on simd processors.
\newblock In \emph{Neural Networks, 1997., International Conference on}, volume~2, pages 933--936. IEEE, 1997.

\bibitem[Banner et~al.(2018)Banner, Hubara, Hoffer, and Soudry]{Scalable8bitBanner2018}
Ron Banner, Itay Hubara, Elad Hoffer, and Daniel Soudry.
\newblock Scalable methods for 8-bit training of neural networks.
\newblock In S.~Bengio, H.~Wallach, H.~Larochelle, K.~Grauman, N.~Cesa-Bianchi, and R.~Garnett, editors, \emph{Advances in Neural Information Processing Systems}, volume~31. Curran Associates, Inc., 2018.

\bibitem[Choi et~al.(2019)Choi, Venkataramani, Srinivasan, Gopalakrishnan, Wang, and Chuang]{Choi2019AccurateAE}
Jungwook Choi, Swagath Venkataramani, Vijayalakshmi Srinivasan, K.~Gopalakrishnan, Zhuo Wang, and Pierce I-Jen Chuang.
\newblock Accurate and efficient 2-bit quantized neural networks.
\newblock In \emph{Conference on Machine Learning and Systems}, 2019.
\newblock URL \url{https://api.semanticscholar.org/CorpusID:96438794}.

\bibitem[Deng et~al.(2009)Deng, Dong, Socher, Li, Li, and Fei-Fei]{Deng2009ImageNet}
Jia Deng, Wei Dong, Richard Socher, Li-Jia Li, Kai Li, and Li~Fei-Fei.
\newblock Imagenet: A large-scale hierarchical image database.
\newblock In \emph{2009 IEEE Conference on Computer Vision and Pattern Recognition}, pages 248--255, 2009.
\newblock \doi{10.1109/CVPR.2009.5206848}.

\bibitem[Dettmers et~al.(2022)Dettmers, Lewis, Belkada, and Zettlemoyer]{Dettmersint8GPT}
Tim Dettmers, Mike Lewis, Younes Belkada, and Luke Zettlemoyer.
\newblock Gpt3.int8(): 8-bit matrix multiplication for transformers at scale.
\newblock In S.~Koyejo, S.~Mohamed, A.~Agarwal, D.~Belgrave, K.~Cho, and A.~Oh, editors, \emph{Advances in Neural Information Processing Systems}, volume~35, pages 30318--30332. Curran Associates, Inc., 2022.

\bibitem[Ghaffari et~al.(2022)Ghaffari, Tahaei, Tayaranian, Asgharian, and Partovi~Nia]{Ghaffari2022IsIA}
Alireza Ghaffari, Marzieh~S. Tahaei, Mohammadreza Tayaranian, Masoud Asgharian, and Vahid Partovi~Nia.
\newblock Is integer arithmetic enough for deep learning training?
\newblock In S.~Koyejo, S.~Mohamed, A.~Agarwal, D.~Belgrave, K.~Cho, and A.~Oh, editors, \emph{Advances in Neural Information Processing Systems}, volume~35, pages 27402--27413. Curran Associates, Inc., 2022.

\bibitem[Gupta et~al.(2015)Gupta, Agrawal, Gopalakrishnan, and Narayanan]{Gupta2015DeepLW}
Suyog Gupta, Ankur Agrawal, Kailash Gopalakrishnan, and Pritish Narayanan.
\newblock Deep learning with limited numerical precision.
\newblock In \emph{Proceedings of the 32nd International Conference on International Conference on Machine Learning - Volume 37}, ICML'15, page 1737–1746. JMLR.org, 2015.

\bibitem[He et~al.(2016)He, Zhang, Ren, and Sun]{He1025ResNet}
Kaiming He, Xiangyu Zhang, Shaoqing Ren, and Jian Sun.
\newblock Deep residual learning for image recognition.
\newblock In \emph{2016 IEEE Conference on Computer Vision and Pattern Recognition (CVPR)}, pages 770--778, 2016.
\newblock \doi{10.1109/CVPR.2016.90}.

\bibitem[Horowitz(2014)]{Horowitz2014Energy}
Mark Horowitz.
\newblock 1.1 computing's energy problem (and what we can do about it).
\newblock In \emph{2014 IEEE International Solid-State Circuits Conference Digest of Technical Papers (ISSCC)}, pages 10--14, 2014.
\newblock \doi{10.1109/ISSCC.2014.6757323}.

\bibitem[Kingsbury and Rayner(1971)]{KiNiRa71}
Nick~G Kingsbury and Peter~JW Rayner.
\newblock Digital filtering using logarithmic arithmetic.
\newblock \emph{Electronics Letters}, 2\penalty0 (7):\penalty0 56--58, 1971.

\bibitem[Kirkpatrick et~al.(1983)Kirkpatrick, Gelatt, and Vecchi]{Kirkpatrick1983SA}
S.~Kirkpatrick, C.~D. Gelatt, and M.~P. Vecchi.
\newblock Optimization by simulated annealing.
\newblock \emph{Science}, 220\penalty0 (4598):\penalty0 671--680, 1983.
\newblock \doi{10.1126/science.220.4598.671}.
\newblock URL \url{https://www.science.org/doi/abs/10.1126/science.220.4598.671}.

\bibitem[Krizhevsky(2009)]{Krizhevsky09learningmultiple}
Alex Krizhevsky.
\newblock Learning multiple layers of features from tiny images.
\newblock Technical report, University of Toronto, 2009.

\bibitem[Le and Yang(2017)]{TinyImageNet2017}
Y.~Le and X.~Yang.
\newblock Tiny imagenet visual recognition challenge.
\newblock \url{http://cs231n.stanford.edu/reports/2015/pdfs/yle_project.pdf}, 2017.

\bibitem[Lee and Edgar(1977)]{LeEdAl77}
Samuel~C. Lee and Albert~D. Edgar.
\newblock The focus number system.
\newblock \emph{IEEE Transactions on Computers}, 26\penalty0 (11):\penalty0 1167--1170, 1977.

\bibitem[Liu et~al.(2023)Liu, Zhang, Zhang, Hao, Du, Hu, Li, and Guo]{liu2023AlsPotq}
Chang Liu, Rui Zhang, Xishan Zhang, Yifan Hao, Zidong Du, Xing Hu, Ling Li, and Qi~Guo.
\newblock Ultra-low precision multiplication-free training for deep neural networks, 2023.

\bibitem[Loshchilov and Hutter(2017)]{Loshchilov2016SGDR}
Ilya Loshchilov and Frank Hutter.
\newblock Sgdr: Stochastic gradient descent with warm restarts.
\newblock In \emph{International Conference on Learning Representations, ICLR}, 2017.

\bibitem[Mitchell(1962)]{Mitchell1962}
John~N. Mitchell.
\newblock Computer multiplication and division using binary logarithms.
\newblock \emph{IRE Transactions on Electronic Computers}, EC-11\penalty0 (4):\penalty0 512--517, 1962.
\newblock \doi{10.1109/TEC.1962.5219391}.

\bibitem[Sanyal et~al.(2020)Sanyal, Beerel, and Chugg]{Sanyal2019NeuralNT}
Arnab Sanyal, Peter~A. Beerel, and Keith~M. Chugg.
\newblock Neural network training with approximate logarithmic computations.
\newblock In \emph{ICASSP 2020 - 2020 IEEE International Conference on Acoustics, Speech and Signal Processing (ICASSP)}, pages 3122--3126, 2020.
\newblock \doi{10.1109/ICASSP40776.2020.9053015}.

\bibitem[Simonyan and Zisserman(2015)]{SimonyanZ14a}
Karen Simonyan and Andrew Zisserman.
\newblock Very deep convolutional networks for large-scale image recognition.
\newblock In Yoshua Bengio and Yann LeCun, editors, \emph{3rd International Conference on Learning Representations, {ICLR} 2015, San Diego, CA, USA, May 7-9, 2015, Conference Track Proceedings}, 2015.
\newblock URL \url{http://arxiv.org/abs/1409.1556}.

\bibitem[Sun et~al.(2020)Sun, Wang, Chen, Ni, Agrawal, Cui, Venkataramani, El~Maghraoui, Srinivasan, and Gopalakrishnan]{Sun20204bit}
Xiao Sun, Naigang Wang, Chia-Yu Chen, Jiamin Ni, Ankur Agrawal, Xiaodong Cui, Swagath Venkataramani, Kaoutar El~Maghraoui, Vijayalakshmi~(Viji) Srinivasan, and Kailash Gopalakrishnan.
\newblock Ultra-low precision 4-bit training of deep neural networks.
\newblock In H.~Larochelle, M.~Ranzato, R.~Hadsell, M.F. Balcan, and H.~Lin, editors, \emph{Advances in Neural Information Processing Systems}, volume~33, pages 1796--1807. Curran Associates, Inc., 2020.

\bibitem[Swartzlander and Alexopoulos(1975)]{SwAl75}
Earl~E Swartzlander and Aristides~G Alexopoulos.
\newblock The sign/logarithm number system.
\newblock \emph{IEEE Transactions on Computers}, 24\penalty0 (12):\penalty0 1238--1242, 1975.

\bibitem[Wang et~al.(2022)Wang, Rasoulinezhad, Leong, and So]{Wang2020NITITI}
Maolin Wang, Seyedramin Rasoulinezhad, Philip H.~W. Leong, and Hayden K.-H. So.
\newblock Niti: Training integer neural networks using integer-only arithmetic.
\newblock \emph{IEEE Transactions on Parallel and Distributed Systems}, 33\penalty0 (11):\penalty0 3249--3261, 2022.
\newblock \doi{10.1109/TPDS.2022.3149787}.

\bibitem[Wu et~al.(2018)Wu, Li, Chen, and Shi]{wu2018WAGE}
Shuang Wu, Guoqi Li, Feng Chen, and Luping Shi.
\newblock Training and inference with integers in deep neural networks.
\newblock In \emph{International Conference on Learning Representations}, 2018.
\newblock URL \url{https://openreview.net/forum?id=HJGXzmspb}.

\bibitem[Zhao et~al.(2022)Zhao, Dai, Venkatesan, Zimmer, Ali, Liu, Khailany, Dally, and Anandkumar]{Zhao2021LNSMadamLT}
Jiawei Zhao, Steve Dai, Rangharajan Venkatesan, Brian Zimmer, Mustafa Ali, Ming-Yu Liu, Brucek Khailany, William~J. Dally, and Anima Anandkumar.
\newblock Lns-madam: Low-precision training in logarithmic number system using multiplicative weight update.
\newblock \emph{IEEE Transactions on Computers}, 71\penalty0 (12):\penalty0 3179--3190, 2022.
\newblock \doi{10.1109/TC.2022.3202747}.

\end{thebibliography}

\end{document}